\documentclass[sn-chicago]{sn-jnl}% Chicago-based Humanities Reference Style
\usepackage{enumerate}
\usepackage{cleveref}
\usepackage{xcolor}
\jyear{2021}%

\theoremstyle{thmstyleone}%
\theoremstyle{thmstyletwo}%

\theoremstyle{thmstylethree}%

\begin{document}

\title[ANNs and the Representation of Human Concepts]{Artificial Neural Nets and the Representation of Human Concepts}

\author{\fnm{Timo} \sur{Freiesleben}}\email{timo.freiesleben@uni-tuebingen.de}

\affil{\orgdiv{Cluster of Excellence: "Machine Learning: New Perspectives for Science"}, \orgname{University of Tübingen}, \orgaddress{\street{Maria-von-Linden-Straße 6}, \city{Tübingen}, \postcode{72076}, \country{Germany}}
%\linebreak
%\newline
%\small{For: \textit{Philosophy of Science for Machine Learning: Core Issues and New Perspectives}, edited by Juan Durán and Giorgia Pozzi}
}

\abstract{
    What do artificial neural networks (ANNs) learn? The machine learning (ML) community shares the narrative that ANNs must develop abstract human concepts to perform complex tasks. Some go even further and believe that these concepts are stored in individual units of the network. Based on current research, I systematically investigate the assumptions underlying this narrative. I conclude that ANNs are indeed capable of performing complex prediction tasks, and that they may learn human and non-human concepts to do so. However, evidence indicates that ANNs do not represent these concepts in individual units. 
    }


\maketitle

\section{Introduction}
\label{sec:intro}

Large parts of the machine learning (ML) community seems to share a remarkable narrative about the representational qualities of Artificial Neural Networks (ANNs) \citep{olah2020zoom,lecun2015deep}:
\begin{enumerate}
    \item ANNs can perform tasks that only humans could perform in the past, such as classifying videos or producing human-like text.
    \item When humans perform these tasks, they rely on \emph{concepts}, that means higher order representations. Humans perceive images as composed of \emph{objects}. They think of texts as following abstract \emph{grammatical rules}.    
    \item To do the same tasks as humans, ANNs must have developed the same concepts as humans.
    \item These concepts are represented in the weights, units, filters, channels, or layers of the ANNs, depending on the model architecture and who you ask.
\end{enumerate}

And indeed, people started searching for evidence in favor of this narrative. They found that early layers in trained Convolutional Neural Nets (CNNs) closely reassemble or even exactly match edge detectors designed by human engineers \citep{olah2017feature}. Later units in CNNs show co-activation with some complex concepts found in images, such as wheels or grass \citep{bau2017network}. Text classifiers have neurons that are connected to the sentiment in texts \citep{radford2017learning} and Large Language Models (LLMs) appear to store factual knowledge largely in mid-layer modules \citep{meng2022locating} and have neurons connected to space and time \citep{gurnee2023language}. 

If the narrative were correct, it would have great practical and philosophical  significance: It would elucidate why ANNs work so well -- they simply develop the right concepts \citep{lecun2015deep}; It would show that human concepts, including our natural kinds \citep{sep-natural-kinds}, indeed arise naturally from efficiently solving complex prediction tasks \citep{watson2023philosophy}; Moreover, it would elucidate how empiricist philosophy can explain human cognition \citep{buckner2018empiricism}.

Yet, I remained skeptical: Aren't neural nets just estimators of conditional densities? Are our concepts really so natural that ANNs have to learn the exact same concepts? Wasn't the point of using ANNs to distribute meaning rather than store it in individual units?

I therefore started my journey to examine the ML literature on concept representation in ANNs. I found that ML researchers remain vague on what they mean by the representation of learned human concepts, and hence they do not provide testable hypotheses, but rather anecdotal evidence \citep{leavitt2020towards}. I have structured my search, and consequently this chapter, along the three key assumptions I found underlying the narrative above. I already give here my conclusions about these assumptions:

\begin{itemize}
    \item \textbf{Assumption I, \emph{ANNs work well}:} Even though the theoretical grounds are still shaky, there is strong empirical evidence that ANNs are solid approximations of the best possible predictors in static data settings (See \Cref{sec:ass1}).
    \item \textbf{Assumption II, \emph{ANNs learn human concepts}:} There is both evidence in favor of this claim (e.g. transfer learning) and evidence against it (e.g. adversarial examples). It is likely that ANNs learn both some human and some non-human concepts (See \Cref{sec:ass2}).    
    \item \textbf{Assumption III, \emph{ANNs represent these learned human concepts by single units}:} Although some evidence could be interpreted in favor of the claim, it turns out to be very weak. Individual units that are supposed to represent concepts often only weakly coactivate with them and do not share their functional role  (See \Cref{sec:ass3}).
\end{itemize}

\paragraph{My goals with this work:}  First, to provide clarity on the representational qualities of current ANNs based on evidence. Second, to provide ML researchers with conceptual guidance on the notion of the \textit{representation of learned human concepts} with the hope of helping them form falsifiable hypotheses and gather further evidence for assumptions I-III.

In this work, I will \emph{not} touch on the question whether ANNs are in any way similar to biological neural nets, nor will I discuss the question of how humans store or process concepts. Readers interested in these questions are pointed to \cite{marblestone2016toward,buckner2018empiricism,bowers2022deep} or \cite{grujivcic2024deep}.

\section{Background}
\label{sec:background}

In this Section, I will first introduce what I mean by the representation of learned human concepts. Second, I will discuss other work concerned with concept representation in ANNs. And third, I will provide a short primer on ANNs.

\subsection{Introducing terms}
\label{subsec:definitions}

What concepts precisely are is widely debated and rarely agreed upon in the literature \citep{sep-scientific-representation}. To avoid entering this minefield, I will focus in this chapter on the widely accepted roles of concepts rather than on their ontology.

\paragraph{Roles of concepts}
Concepts take an \emph{abstractional} role \citep{yee2019abstraction}. To some objects in the world a concept applies, to others it does not \citep{peacocke1992study}. Take the concept of a \emph{metro}. Both the Munich and the London metro are instantiations of the concept metro. Having learned a concept also means being able to decide (relatively) competently whether a concept applies to an object or not.\footnote{There are two ways to define the abstractional role of concepts -- extensionally and intensionally. \emph{Extensional:} Definition in terms of all the objects or ideas to which the concept applies. I can define \emph{metro} by pointing to all metro systems on earth. \emph{Intensional:} Definition by the properties that an object or idea must have for the concept to apply. One can define what a tree is by specifying a set of conditions for being a tree, e.g. having a stem, growing roots, being made out of wood, etc. This requires further, more basic concepts.}

Concepts take a \emph{functional} role \citep{lalumera2010concepts}. Having learned a concept means being able to perform certain tasks and entertain certain properties \citep{keil1992concepts}. For example, a kid has learned the concept of addition if she can give the correct answer in a series of moderately difficult addition problems she has never faced before. Clearly, learning a concept comes in degrees, one may have learned a concept more or less well. The kid may be able to perform addition when positive numbers are involved, but not when one of the numbers is negative.

Other roles could be pushed here, like the compositionality of concepts \citep{del2016prototypes,hampton2012typicality}. We can, for instance, in some contexts define a metro line via its stops and the stops via their locations. For this chapter, I restrain myself to the abstractional and functional role, which I both consider as essential roles of concepts. It is extremely difficult and perhaps impossible to give necessary and sufficient conditions for the concept \emph{concept}, so I will stick to these agreeable family resemblances \citep{wittgenstein2010philosophical,hampton2006concepts}.

\paragraph{Human concepts}
A concept is a \emph{human concept}\footnote{Animals also learn concepts, both concepts similar to humans and some animal specific \citep{andrews2017routledge}.} if there is at least one human being who uses the concept and can  communicate effectively about it. Metro is a human concept as I and probably other human beings can use the concept and communicate about it. 

Indeed, what counts as a human concept is constantly evolving \citep{keil1992concepts}. New concepts emerge as new objects and new goals are developed by humans. One hundred years ago \emph{smartphone} was not a human concept, today every kid learns this concept. Similarly, 100,000 years ago, people probably had rich conceptual frameworks related to hunting that we no longer understand.

In the philosophical literature there is the idea of natural kinds -- i.e. concepts that exist in the world independently of human goals \citep{sep-natural-kinds}. These concepts are considered particularly suitable for drawing scientific inferences, examples of supposedly natural kinds are the elements in the periodic table or elementary particles in physics. \cite{quine1969ontological} believed that humans have an innate capacity to judge similarities between objects and find natural kinds. If ANN models develop the same kind of concepts as we do, this would suggest that humans also develop concepts to solve complex prediction tasks, rather than having an innate ability to find natural kinds \citep{williams2018predictive}.

\paragraph{Representing learned concepts in a model}
A prediction model represents\footnote{Perfect representation is clearly the exception, much more often models only approximately represent [See Chapter 12 in this volume].} a learned concept $C$ in model part $U$, if 
\begin{enumerate}[i)]
    \item \emph{Coactivation: }the model part $U$ coactivates when exhibited to instances of the concept $C$.
    \item \emph{Functional role: }removing or manipulating the model part $U$ takes away or changes the functional role of the concept $C$ in the prediction.
\end{enumerate}

Model parts can be variables, parameters, mathematical relations, or compositions thereof. As should be evident, representation is meant here as in scientific representations rather than mental representations \citep{sep-scientific-representation,duede2023representational}. Newton's laws of gravity represent the concept of mass in the variable $m$, since an increase in the mass of an object leads to the prediction of an increased gravitational pull on other bodies. 
An ANN represents the concept of addition in a model part $U$ when each time it performs an addition, the part is activated and removing the unit results in ANNs no longer being able to perform common addition problems that they were previously capable of solving.

\subsection{Related work}
\label{subsec:related}
There are both, philosophical papers and technical papers on ML that touch upon the question of what ANNs learn and represent.

\paragraph{Proponents that ANNs represent concepts in individual units}
\cite{olah2020zoom} are maybe the strongest proponents of Assumption III. Based on examining many neurons via activation maximization, they conclude that units in ANNs store individual disentangled concepts and that the algebraic operations of neurons can be seen as performing logical reasoning with these concepts. They moreover push the claim that there is a universal set of concepts that must be developed to perform certain tasks. Similarly, \cite{bau2017network} believe that networks can ultimately be dissected into hierarchies of individual concepts, from shapes and colors to scenes and objects. \cite{raz2023methods} discusses both their methods (I will introduce these methods in more detail in \Cref{sec:ass3}), namely network dissection \citep{bau2017network} and activation maximization \citep{olah2017feature}, and moreover testing with concept activation vectors (TCAV) by \cite{kim2018interpretability}. He believes that the methods provide compelling evidence for the representation of concepts in individual units, but argues for the demand of additional methods and applying a variety of methods for identifying concepts in ANNs.

I will argue why we should be skeptical of the claim that ANNs store learned human concepts in individual units, in particular I will question how conclusive the evidence obtained from the above methods really is.

\paragraph{Skeptics if ANNs represent concepts in individual units}

\cite{freiesleben2022scientific} argue against the view that ANNs represent human concepts in individual units and instead argue that ANNs only represent conditional probabilities. Both options will be discussed below, together with the third option that ANNs learn human concepts but store them holistically. \cite{duede2023representational} argues that the confusion about representation in ANNs arises from mixing functional and relational conceptions of representation. I provide further arguments why it is questionable to see ANNs as representing concepts in individual network units. \cite{leavitt2020towards} argue that current research on concepts in ANNs is too driven by intuitions rather than testable hypotheses and advocate for understanding ANNs as distributed representations. Similarly, Kieval emphasizes that while ANNs can learn useful higher-order representations, these are useful for the ML model to perform the task at hand, and not for the human [See Chapter 10 in this volume]. He points out that ANNs rely on distributed rather than a unit-wise representation, and argues against representationalist approaches to ANNs overall. I will go into more detail on why at least the representation of concepts in individual units is difficult to defend.

\paragraph{Other philosophical positions on concepts in ANNs}

\cite{buckner2018empiricism} argues that a specific form of ANNs, namely convolutional neural networks (CNNs), can help understand the mechanisms in the brain that allow us to learn abstractions from examples and generate examples from abstractions, and in turn solve philosophical problems of empiricist philosophers. However, Buckner lays no claims to whether these abstractions in ANNs are stored in individual units or whether ANNs learn human concepts. 
Similarly, \cite{lopez2021throwing} argues that CNNs and generative adversarial networks learn emergent natural object categories from images without human supervision. But like \cite{buckner2018empiricism}, he does not address the question of whether these categories are proper concepts that fulfill ``objective criteria of correctness and intersubjective agreement'' \citep[p.10024]{lopez2021throwing}. \cite{lopez2021throwing} bases his arguments on the empirical work by \cite{bau2017network} and \cite{bau2018gan}, but notes more clearly the limitation of their approach due to human bias in judging the semantics of a unit and the sometimes quite distributed representation of categories.

Work by \cite{ilyas2019adversarial} indicates that ANNs learn concepts that humans understand as well as concepts humans do not use, like certain predictive texture structures in images. Based on these findings, \cite{buckner2020understanding} discusses how examining predictive concepts that humans do not yet know can shed new light on the problem of natural kinds. \cite{boge2023functional} argues that ANNs do not learn concepts, but rather develop what he calls \emph{functional concept proxies} -- representations that take the inferential role of specific concepts in certain contexts, but do not capture their semantics and their relations to other concepts. However, like \cite{buckner2020understanding}, he beliefs, based on \cite{ilyas2019adversarial}, that ANNs can develop concept proxies for meaningful scientific concepts that humans have not learned and may not have access to.

\subsection{A short primer on ANNs}

Artificial neural networks (ANNs) consist of primitive stacked units called neurons \citep{goodfellow2016deep}. Each neuron is composed of two parts: a weighted linear sum and an activation function. The architecture of ANNs is determined by the number of parallel units, the layers in which they are stacked, and the connectivity between layers.

The so-called ``universal approximation theorem'' shows that these simple stacked units are extremely expressive. In fact, any continuous function can be arbitrarily well approximated with an ANN of depth two \citep{hornik1989multilayer}.\footnote{While in theory one hidden layer with high width is enough to learn arbitrary functions, in practice depth is crucial. Research suggests that this might be connected to the greater expressivity of deeper networks in finite parameter regimes \citep{lu2017expressive,rolnick2017power}.} However, there is a significant gap between theoretical existence  and the practical search for such functions. We need so-called learning algorithms to find the right neural networks that approximate the function of interest. These algorithms use the error that models make in their predictions to update the weights in the linear sums. This is done using methods like \emph{gradient descent}, which move the weights in the direction of their gradient in the hope that fewer prediction errors will occur.

How one initializes the weights and which architecture one chooses has a large impact on how successful the search will be. These choices are usually referred to as \emph{inductive biases}.  The two architectures most commonly discussed in the literature are: First, convolutional neural networks (CNNs) \citep{lecun2015deep}, where only neighboring nodes are connected. This architecture has been designed for image recognition tasks. Second, the transformer architecture \citep{vaswani2017attention}, which relies on an attention mechanism to better store contextual information. This architecture was originally developed for language translation, but is now successful in all kinds of setups and forms the core of current chatbots, like ChatGPT.

The ANN obtained from the learning algorithm, that is after the learning process, is called the \emph{prediction model}. In this chapter, when I talk about prediction models, I refer to these trained models and investigate whether they perform well and ideally develop human concepts. The focus will be on supervised ML models that perform prediction tasks after being fed labeled data.

\section{What ANNs do learn: conditional probabilities}
\label{sec:ass1}
Let's check Assumption I: ANNs work well. What does it mean for an ANN to work well? In supervised learning, the goal of ANNs is to predict a given target feature $Y$ (e.g. the correct animal) from a set of predictors $X_1,\dots,X_p$ (e.g. pixels from image of animal) as accurately as possible. What constitutes a good prediction depends on how we define the so-called loss function $L$. Categorizing a leopard as a leopard seems uncontroversial, but what about errors? One may argue that mistaking a leopard for a cheetah is less bad than mistaking it for an elephant. But one could also hold that any misclassification is equally bad.

The loss function determines the best possible prediction functions from $X$ to $Y$, also called the (Bayes) optimal predictor \citep{hastie2009elements,von2011statistical}.\footnote{Note however, that the best predictor does not determine a single best function $f:\mathcal{X}\rightarrow \mathcal{Y}$. It only determines the functional values for $\mathbb{P}(X=x)>0$ or in continuous cases, where the corresponding density is positive. For example, some adversarial attacks rely on probing the model in regions with zero density, where the model is not necessarily competent \citep{szegedy2013intriguing} and must extrapolate \citep{freiesleben2023beyond}.} For example, if we take mean squared error (MSE) as our loss in a regression task, the optimal predictor is the conditional expectation $\mathbb{E}[Y\mid X]$. For the so-called 0-1 loss that assigns the same error of 1 to every misclassification, the optimal predictor is the $\underset{y\in\mathcal{Y}}{\text{arg max}}\; \mathbb{P}(Y=y\mid X)$. In general, the best predictor is a quantity that can be derived from $\mathbb{P}(Y\mid X)$.\footnote{Since people are often confused when it comes to causality: this already shows that current ANNs neither learn nor represent the causal mechanism between $X$ and $Y$ [See Chapter 6 in this volume]. They learn a single distribution, rather than a set of distributions. ANNs completely ignore causal dependencies between the variables $X_1,\dots,X_n$ and exploit effects of the target $Y$ or spurious correlations just as much as causes \citep{konig2023improvement}.}

\subsection{Theoretical guarantees are lacking}
Can we prove that ANNs learn optimal predictors, or at least good approximations of them? In statistical learning theory, originally the main framework for studying such questions, there exists a trade-off between the learning guarantees we get and the expressivity of the function class  \citep{vapnik1999nature}. As we already discussed above, ANNs are extremely expressive, they have infinite VC-dimension. Unfortunately, this implies that we cannot obtain learning guarantees from statistical learning theory for ANNs by any of the standard theorems \citep{sterkenburg2023statistical}.

Often in the training process of ANNs, we have more model parameters than training data. Hence, there are infinite models that perfectly approximate the training data but do not approximate the optimal predictor -- they overfit the data \citep{hastie2009elements}. The double-descent phenomenon shows that ANNs are different again, here more parameters can increase the generalization qualities of ANNS \citep{belkin2019reconciling}. \cite{belkin2021fit} speculated that this quality may be related to the greater ability to select the model that minimizes structural risk among all interpolating models.

It is not even theoretically clear why ANNs trained with gradient based approaches learn models that perfectly interpolate the training data. Since the surface of the loss is non-convex, there are many local minima where the optimization could get stuck \citep{kingma2014adam}. \cite{belkin2021fit} again provides an idea for why local minima are not a problem: in overparametrized regimes, every local minimum is a global minimum, since there is a global optimum in close proximity around every weight vector.

\subsection{Practical success is evident}

Empirically, we have seen the vast success of ANNs: 

\begin{itemize}
    \item They have been successful for all kinds of data, from tabular to image and geo-spatial to text and speech data \citep{erickson2017machine,oikarinen2019deep,ren2021deep}.
    \item Moreover, ANNs have been successfully used in all kinds of domains, e.g. optimizing industrial processes, providing more deceiving advertisement, or diagnosing patients \citep{yang2020using,zhang2019deep,gao2020machine}.
    \item ANNs learn even highly complex interactions efficiently and usually improve with bigger data even of poor quality, they can even perfectly fit synthetically produced random noise \citep{zhang2021understanding}.
\end{itemize}

The empirical success of ANNs cannot be denied, suggesting that ANNs have theoretical properties yet to be revealed. However, there are also cases where ANNs fail even in static distributions, namely in the presence of shortcuts \citep{geirhos2020shortcut,freiesleben2023beyond}; ANNs may learn to distinguish huskies from wolfs based on the snow in the background rather than the appearance of the animals \citep{ribeiro2016should}. Such mistakes can often be avoided by augmenting data or choosing data representational for the task \citep{kim2021biaswap,freiesleben2023beyond}.

\paragraph{Taking stock} Theoretical guarantees are indeed lacking. But I believe the empirical success indicates that ANNs approximate optimal predictors well in many settings.

\section{What ANNs might learn: human concepts}
\label{sec:ass2}
Let's look into Assumption II: ANNs learn human concepts.
It could be argued that an understanding of human concepts is necessary to perform complex classification tasks - hence ANNs learn these too. However, there are many different features on which classification can be based. Imagine a network that distinguishes sweaters from shoes, there can be different strategies:
\begin{itemize}
    \item \emph{Definitional features:} using features such as the shape of the objects that define the class.
    \item \emph{Likely features humans use:} using features that humans also use that are likely cues for the target class, such as the pleats of the sweater or shoelaces.
    \item \emph{Likely features humans do not use:} using features that humans do not use that are likely cures for the target class, such as sweater typical texture structures.
    \item \emph{Spurious features:} using features that are specific to the data in the dataset. They are neither necessary nor likely cues for the class, think of certain colors, fabrics or background conditions.
\end{itemize} 

Which of these do ANNs learn?

\subsection{Transfer learning indicates that reliable and general concepts are learned}
\begin{figure}[h]
\centering
  \includegraphics[width=0.85\linewidth]{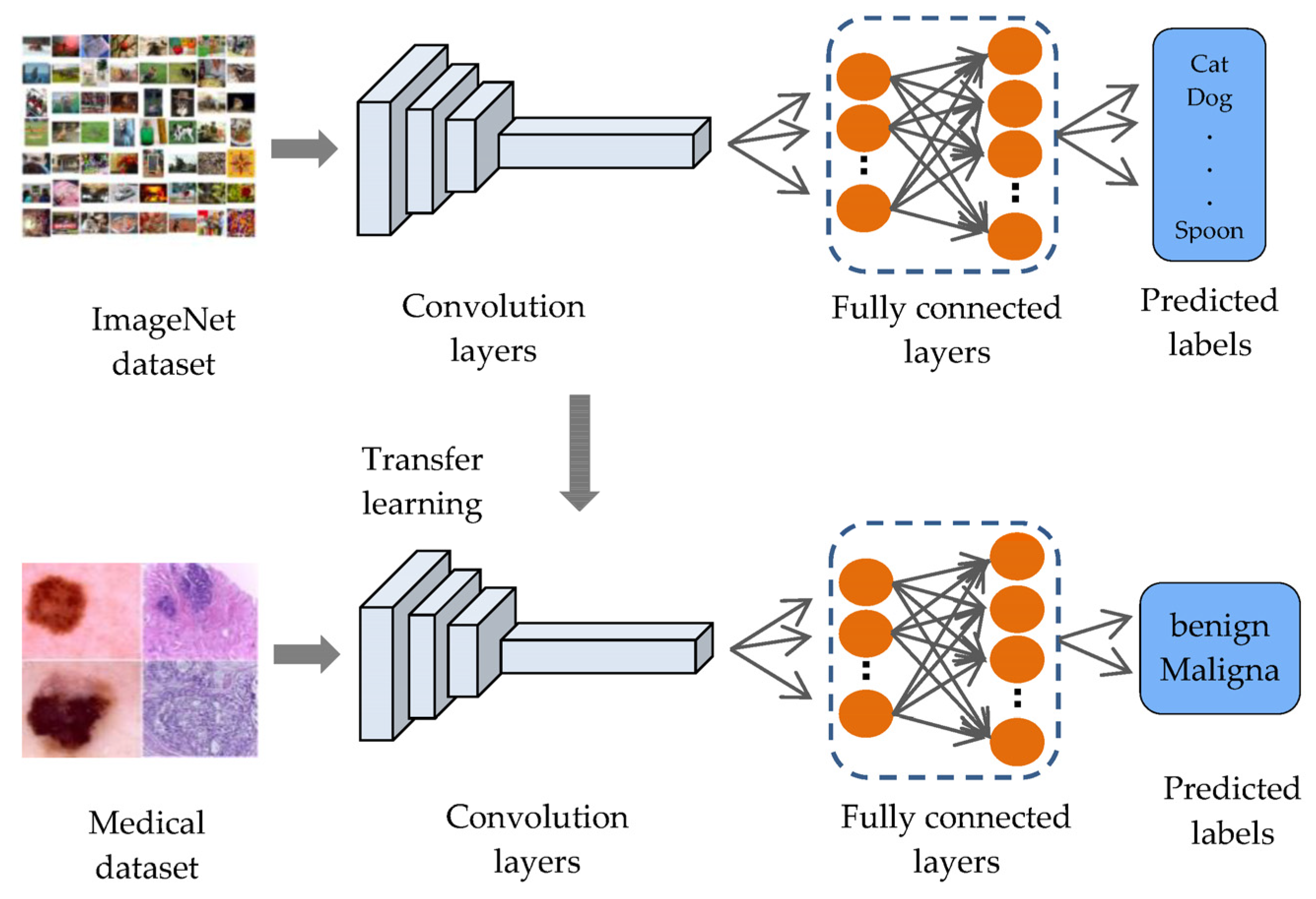}
  \caption{An illustration of transfer learning by \cite{mukhlif2023incorporating}. In this case, filters of a CNN are transferred from one to another task. The fact that this strategy works means that models have learned general and reliable concepts.}
  \label{fig:transfer}
\end{figure}

Transfer learning means that parts of models can be reused to solve related but different tasks \citep{tan2018survey}, as shown in \Cref{fig:transfer}. To perform a new image classification task, instead of learning a completely new model, it is common to use well-trained models such as ImageNet or ResNet and fine-tune them to the task at hand \citep{kornblith2019better,shermin2019enhanced}. Similarly, for a new text generation task, one would fine-tune existing language models like ChatGPT rather than starting from scratch \citep{zhong2023can}. This common practice produces excellent results, even with little training data.

Transfer learning shows that ANNs do not only learn spurious features, otherwise the learned features would not generalize to other tasks. Instead, the features must capture robust pattern in the data that are useful across problems and reliable indicators to make accurate predictions.

\subsection{TCAV indicates that ANNs learn human concepts}
Still ANNs might only learn concepts humans do not use. The interpretability technique \emph{testing with concept activation vectors (TCAV)} by \cite{kim2018interpretability} provides evidence that some human concepts are learned.

Intuitively, TCAV measures how (marginally) adding or subtracting concepts from instances changes the local or global behavior of the model. It does so via a few steps:
\begin{enumerate}[i]
    \item To use the technique, you need a trained ANN image classification model and a dataset where to some instances a certain concept applies and to others not.
    \item Run each instance of the above dataset through your trained classifier and store the activations of layers of the ANN in a table. Then, use a simple logistic regression model to separate images that instantiate the concept from those images who don't based on the activations stored in the table.
    \item The orthogonal of the linear curve that separates the concept class from the non-concept class defines your concept vector. A concept influences the behavior of your model if adding or subtracting the concept vector (multiplied by some scalar) changes the predictions of single or many instances.
\end{enumerate}

\cite{kim2018interpretability} show for instance that ImageNet relies on colors like red, yellow, and blue to classify fire engines; Inception3 shows problematic behavior by relying on gender to classify aprons, specifically associating them with women.

Note that the concepts in TCAV are ultimately defined by the objects that belong to a concept class. This may be problematic as multiple concepts can be present in this class \citep{raz2023methods}. For example, we want to get the concept vector for the concept \emph{basketball}. However, it could be that basketballs are the only orange objects in the concept dataset, in which case our concept vector could represent the concept \emph{orange} and not the concept \emph{basketball}. Also, the TCAV method cannot actually be applied to learn which non-human concepts the model relies on, as this would require giving instances of these concepts.

\subsection{Adversarial examples show that models may rely on other concepts than humans}

\begin{figure}[h]
\centering
  \includegraphics[width=0.85\linewidth]{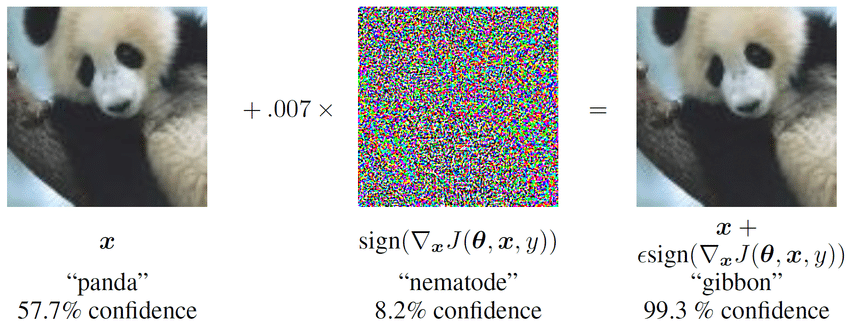}
  \caption{One illustration of an adversarial example by \cite{goodfellow2014explaining}. An image classification model misclassifies a panda as a gibbon after adding noise on the image. This indicates that the model relies in its prediction on some features that humans do not rely on.}
  \label{fig:adversarial}
\end{figure}

Adversarial examples describe cases where slight manipulations of data lead to incorrect predictions:
\begin{itemize}
    \item Buying a bird can make you eligible for a loan \citep{ballet2019imperceptible}. 
    \item As illustrated in \Cref{fig:adversarial}, adding imperceptible noise makes image classifiers believe that a panda is a gibbon \citep{goodfellow2014explaining}. 
    \item A small printed patch makes models classify every object next to it as a toaster \citep{brown2017adversarial}.
    \item Deceiving prompts can lead large language models to hallucinate \citep{mckenna2023sources}.
\end{itemize}

If ANNs would have learned all and only human concepts, such attacks would not be possible. One may take adversarial examples therefore as evidence that ANNs learn mostly faulty concepts. Recently, however, a more nuanced viewpoint has been put forward by \cite{ilyas2019adversarial}: well trained ANNs indeed learn reliable concepts that generalize from the training to the test set and often also to other natural instances. Even though those concepts are reliably predictive of the target class in natural settings, humans do not use these concepts. For example, a sweater typical texture is very predictive of sweaters, but humans do not rely on it, probably because other features are even more predictive and easier to recognize from greater distances.

\cite{ilyas2019adversarial} concluded that ANNs learn both, human and non-human concepts. It is an open question what the nature of these non-human concepts is and whether humans can in principle understand them \citep{buckner2020understanding}.  \cite{bowers2022deep} present interesting ideas on this question: ANNs are more likely to pay attention to complex textures than to shape, and even when they do pay attention to shape, it is more likely to be local than global shape \citep{geirhos2018imagenet}. Another approach to better understand these non-human concepts could be via interpretability techniques: saliency maps can highlight the regions of the adversarial input that most influenced the model's decision \citep{wang2022adversarial} and the distribution of TCAVs of adversarial examples differs from the distribution of TCAVs of normal instances \citep{kim2018interpretability}.

To me, adversarial examples clearly show that models often do neither learn the hierarchical importance of features nor their causal role
\begin{itemize}
    \item \emph{Hierarchical importance:} Even if a certain kind of texture on a panda appears to a classifier as very gibbonish, this should be overruled by the shape of the animal that is indeed very distinct between a panda and a gibbon.
    \item \emph{Causal role:} Even if people with more pets are statistically more likely to pay back their loans, this does not mean that increasing the number of pets would make them more creditworthy \citep{ballet2019imperceptible,freiesleben2022intriguing}. 
    \end{itemize}

\paragraph{Taking stock}
We can conclude that ANNs may use all kinds of features to perform their predictions, those that are definitional, predictive features used and not used by humans, and sometimes also spuriously correlated features. There is more than one way to bake a cake. I would speculate that from a purely predictive perspective there are probably many more useful concepts humans do not use than there are concepts humans do use.

\section{What ANNs probably don't do: represent human concepts in units}
\label{sec:ass3}

Let us turn to Assumption III: ANNs represent learned human concepts by single units. I have argued so far that ANNs can learn concepts and that they may even learn human concepts, but where do they store them? There is the widespread belief that ANNs store all concepts individually in single units of the network \citep{olah2020zoom,bau2017network}. What is the evidence on that?

\subsection{Activation maximization can be misleading}
\begin{figure}[h]
\centering

   \includegraphics[width=0.85\linewidth]{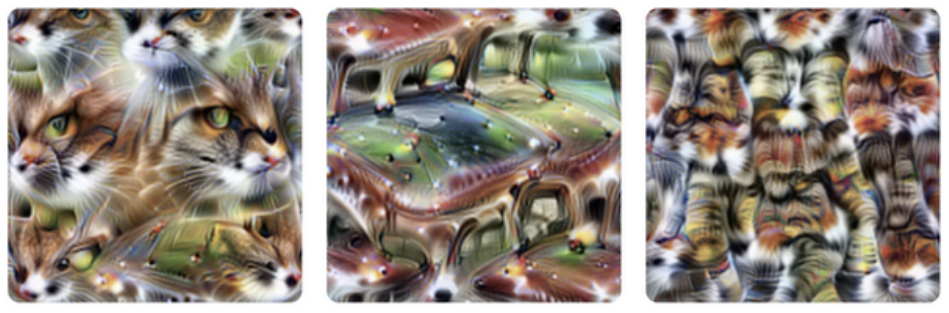}

    \caption{\cite{olah2017feature,olah2020zoom} applied activation maximization three times to one single unit of an ANN with slightly different initializations and objectives. Humans may find some semantically meaningful structures in the images (e.g. cat heads, car fronts, and bee bodies), but this only works for some units and can be conflicting as in this case. \cite{olah2017feature} calls this neuron polysemantic, but how can we be sure that even there is any semantics based on these non-representative examples?}
  \label{fig:actMax}
\end{figure}

One technique that got the narrative of representation in single units thriving was \emph{activation maximization} \citep{olah2017feature}. It asks the following question: What input would maximally trigger an individual neuron/unit in a neural network? This problem can be solved by gradient descent, however, this time the parameters we optimize over are not the weights of the network but the input features. Applied to image classifiers, activation maximization generates fancy images as shown in \Cref{fig:actMax}.

What can be deduced from such images? They require human interpretation. In many of such images, humans see certain concepts present to a greater or lesser degree. If one concept is dominant in an image according to human judgment, the neuron is assumed to represent the concept.

There are many reasons to be skeptical about such conclusions:
\begin{itemize}
    \item \emph{Representativeness of individual image:} We should not take the image we get as representative of the neuron. The image varies greatly depending on the image initialization, the designed objective, and the random seeds. These choices have a large impact on the generated images and result in very different detected concepts (See  \Cref{fig:actMax}). Also, minimizing activation rather than maximizing activation again points to different concepts \citep{olah2017feature}.     
    \item \emph{Human biases:} Ultimately, humans are interpreting the images generated from activation maximization, which indeed introduces a bias \citep{raz2023methods,lopez2021throwing}. For example, there is a strong confirmation bias towards finding human concepts. As we know from adversarial examples, humans are probably incapable of recognizing other (more texture based) concepts that might as well be present in the images. More generally, the images presented do not look natural and invite ambiguity -- ideal conditions for leading people to be selective about what they see or fantasize \citep{leavitt2020towards}.
    \item \emph{Functional role derived from single image:} Most critically, neurons, even when associated with a concept, do not necessarily take the functional role of the concept in the prediction. Activation maximization as a technique sheds no light on the functional role of neurons. We know from other cases that removing neurons that are seemingly meaningful concepts has little to no effect on the models performance \citep{donnelly2019interpretability}. Is the classifier getting worse in classifying cats if the cat heads unit is missing?
\end{itemize}

\subsection{Network dissection is impressive in its story but not in its details}
A more systematic approach for concept detection than activation maximization was taken by \cite{bau2017network}. Their framework, called \emph{network dissection}, aims to decompose image classifiers into disentangled components with semantic meaning. Intuitively, they test whether the region in images where humans see concepts matches the representations of those images in specific channels of the neural network.

What their method requires is a standard labeled dataset for which there are additionally labeled concepts, i.e. the concepts humans see in the images are highlighted in the images. Next, one has to decide on a particular channel in the image classifier that should be matched with a certain concept. The main idea is then to insert images into the image classifier and look at the activations of the channel under consideration. The extent to which the top activations of the channel align with the concept region highlighted by a human annotator is a measure for how well the channel represents the concept. Formally speaking, we take the intersection of the channel activation and the labeled concept region and divide it by their union, which gives us the so-called intersection over union (IoU) score.

\cite{bau2017network} differentiate between color, texture, material, scene, part, and object concepts and find for each type coactivating units in the neural network. They show that the alignment between concepts and units is not random, namely if rotations are performed on the channels, the alignment decreases. Moreover, they find that deeper and wider networks show a greater number of concepts stored in single units. 

The work seems indeed impressive. However, looking into the details, I have found great weaknesses: 
\begin{itemize}
    \item The numbers they present are really not impressive. Even the best coactivating example units they show only have a maximum IoU of 30\%. This means human concept regions and unit activations match only to 30\%. Furthermore, the threshold for a unit representing a concept is set surprisingly low, with an IoU score of 0.04 without further explication.
    \item Even for ResNet, the most ``interpretable network''  according to the authors, only a small fraction of units has an IoU above 4\% with some concept. Conversely, only about a quarter of the concepts in the concept dataset can be matched to any of the units. At the same time, most concepts can be linked to many different channels rather than just one; In ResNet, up to 37 units can be linked to the concept ``airplane''.    
    \item As with activation maximization, network dissection does not reveal whether the units actually play the functional role of the concepts, so whether their removal would affect the network functioning with respect to these concepts. To make network dissection useful for interpretability, we would need to understand mechanistically what role these units take in the prediction. Fortunately, this question was explored in their follow-up work \cite{zhou2018revisiting}. But again, the results seem unconvincing to me: removing units has little effect on overall performance, and a relatively small effect on class performance (on average around 8\% performance loss, but admittedly of 50\% in one extreme case). Thus, removing all 37 airplane units likely results in only a minor reduction in ResNet's ability to detect airplanes. This is evidence against the units representing this concept and, methodologically, it is another call for ablation studies when it comes to concepts in ANNs. 
\end{itemize}

\subsection{Selectivity is not necessary for better performance}

Where does this leave us? What \cite{bau2017network}  has shown, at least, is that -- some units in some networks respond systematically differently for objects of some classes compared to objects of other classes. This behavior is called ``selectivity''. But is selectivity somehow advantageous for the task at hand? Lead selectivity favoring architectures or training methods to better predictive performance?

Already \cite{bau2017network} noted that selectivity is not a prerequisite for predictive performance. In fact, many widely used techniques such as self-supervision, dropout or batch normalization, have been found to reduce selectivity but not performance \citep{morcos2018importance}. \cite{leavitt2020selectivity} approach the question of how selectivity and performance relate to each other systematically by introducing a regularization term that controls selectivity. The term allows selectivity to be both enforced and impeded in training. They show that impeding class selectivity can even improve performance, conversely enforcing class selectivity impedes performance. In a follow-up study they show that impeding selectivity increases robustness to natural changes, while enforcing selectivity can lead to improved adversarial robustness to gradient based attacks \citep{leavitt2020relationship}.

\paragraph{Taking stock}

I share \cite{leavitt2020towards} opinion: ``One’s skepticism should be proportional to the feeling of intuitiveness.'' It needs more than single fancy images to call a unit in an ANN a concept. Representing a concept also goes beyond coactivation or highlighting the same regions in 4\% of cases. To make the idea of human concepts in ANNs really worthwhile to the interpretability project, we need to systematically understand when human concepts emerge and check if they represent in the sense of coactivating AND taking the functional role of concepts.

\section{Discussion}

In this chapter, I examined the remarkable narrative that ANNs learn human concepts and store these concepts in individual units. I differentiated between three claims that underlie the narrative: 1. ANNs work well; 2. ANNs learn the concepts humans learn; 3. ANNs store these concepts in individual units. I found solid evidence for the first, mixed evidence for the second and questionable evidence on the third claim. I hope the arguments that I have presented will lead to some skepticism against the narrative. Further evidence must be gathered to give a decisive answer. I hope that my two criteria for concept representation -- coactivation and functional role -- can guide researchers in forming testable hypotheses on the questions.

\paragraph{Concepts beyond supervised learning}
The view that I described in this chapter focused mostly on the representation of concepts emerging in ANNs through supervised learning. The reason is that many techniques that investigate representations focus on ANNs trained with a supervised learning approach. I believe that unsupervised learning techniques and particularly generative models show greater promise in representing concepts, as generating requires greater concept ontologies than classifying. GAN-dissection by \cite{bau2018gan} is a strong step in the right direction as it is studying the functional role of units under manipulations. Similarly, causal representation learning aims to enforce the emergence of semantically meaningful concepts by posing additional constraints \citep{scholkopf2021toward}.

\paragraph{Enforcing the emergence of human concepts}
In some cases, it might be desirable to enforce ANNs to rely on the same features as humans. Think of safety or robustness issues like in autonomous driving. Also, features that humans rely on are features that humans understand -- interpretability is and will be a key concern when it comes to interacting with ML systems \citep{lipton2018mythos}.

Can the representation of human concepts in ANNs possibly be enforced? Concept bottleneck models provide one example of how this might work if we incorporate human supervision \citep{koh2020concept}. However they require a labeled concept dataset that is not normally available. I believe that if representing disentangled concepts in individual units is really what we are after, as a first step (as \cite{leavitt2020towards} has done) we should investigate training methods that enforce selectivity without sacrificing performance. However, this might be hard.

\paragraph{Natural kinds}
Even if they cannot store them in single units, ANNs seem to develop reliable concepts. It would be interesting to find out how to make ANNs learn human concepts or even scientific natural kinds (given they exist). This may require additional tasks beyond prediction such as generating data or acting in a real-world environment. Such an approach, where additional tasks are imposed, might ultimately allow us to characterize how human concepts emerge and what constitutes natural kinds. But perhaps such a characterization is impossible. Human concepts may not capture natural kinds in the world or be the result of performing a particular task, but rather they may just be conventions of social entities.

\paragraph{Dangers of the narrative}
I get the impression that the ML community is too convinced of the narrative to really drop it. People want to believe that ANNs are logical concept machines that only need to be scrutinized to become fully transparent. I think it is dangerous for this narrative to go unquestioned. It leads to research that is prone to confirmation bias, and that bias has real consequences for users. Some interpretability methods already build explanations to end-users based on the assumption that units represent human concepts, like for instance \cite{achtibat2023attribution}. Faulty assumptions will lead to faulty explanations. 

The questions that I think researchers should ask are: Is it desirable that ANNs store human concepts in individual neurons? If yes, what learning methods allow us to enforce this behavior? If not, how can we better investigate the distributed representation of concepts? 

To advance interpretability as a field, we must start forming proper falsifiable scientific hypotheses in the Popperian sense \citep{popper2005logic} rather than generating narratives based on fancy images and human biases. My hope is that my notion of human concept representation is seen as a challenge to ML researchers -- show me a network where individual units coactivate with human concepts and removing them destroys the functional role of the concepts in prediction and I am sold that this network represents human concepts.

\newpage

\bmhead{Conflict of interest}

I declare no conflicts of interest.
         
\bmhead{Acknowledgments}
I thank Giuseppe Primiero and Juan Duran for feedback on earlier versions of this manuscript. Also, I acknowledge support by the Carl Zeiss Foundation (Project: Certification and Foundations of Safe Machine Learning Systems in Healthcare).

\newpage

\bibliography{sn-bibliography}

\begin{thebibliography}{}
\providecommand{\doi}[1]{\url{https://doi.org/#1}}
\bibcommenthead

\bibitem[\protect\citeauthoryear{Achtibat, Dreyer, Eisenbraun, Bosse, Wiegand,
  Samek, and Lapuschkin}{Achtibat et~al.}{2023}]{achtibat2023attribution}
Achtibat, R., M.~Dreyer, I.~Eisenbraun, S.~Bosse, T.~Wiegand, W.~Samek, and
  S.~Lapuschkin. 2023.
\newblock From attribution maps to human-understandable explanations through
  concept relevance propagation.
\newblock {\em Nature Machine Intelligence\/}~{\em 5\/}(9): 1006--1019 .

\bibitem[\protect\citeauthoryear{Andrews and Beck}{Andrews and
  Beck}{2017}]{andrews2017routledge}
Andrews, K. and J.~Beck. 2017.
\newblock {\em The Routledge handbook of philosophy of animal minds}.
\newblock Taylor \& Francis.

\bibitem[\protect\citeauthoryear{Ballet, Renard, Aigrain, Laugel, Frossard, and
  Detyniecki}{Ballet et~al.}{2019}]{ballet2019imperceptible}
Ballet, V., X.~Renard, J.~Aigrain, T.~Laugel, P.~Frossard, and M.~Detyniecki.
  2019.
\newblock Imperceptible adversarial attacks on tabular data.
\newblock {\em arXiv preprint arXiv:1911.03274\/} .

\bibitem[\protect\citeauthoryear{Bau, Zhou, Khosla, Oliva, and Torralba}{Bau
  et~al.}{2017}]{bau2017network}
Bau, D., B.~Zhou, A.~Khosla, A.~Oliva, and A.~Torralba 2017.
\newblock Network dissection: Quantifying interpretability of deep visual
  representations.
\newblock In {\em Proceedings of the IEEE conference on computer vision and
  pattern recognition}, pp.\  6541--6549.

\bibitem[\protect\citeauthoryear{Bau, Zhu, Strobelt, Zhou, Tenenbaum, Freeman,
  and Torralba}{Bau et~al.}{2018}]{bau2018gan}
Bau, D., J.Y. Zhu, H.~Strobelt, B.~Zhou, J.B. Tenenbaum, W.T. Freeman, and
  A.~Torralba 2018.
\newblock Gan dissection: Visualizing and understanding generative adversarial
  networks.
\newblock In {\em International Conference on Learning Representations}.

\bibitem[\protect\citeauthoryear{Belkin}{Belkin}{2021}]{belkin2021fit}
Belkin, M. 2021.
\newblock Fit without fear: remarkable mathematical phenomena of deep learning
  through the prism of interpolation.
\newblock {\em Acta Numerica\/}~30: 203--248 .

\bibitem[\protect\citeauthoryear{Belkin, Hsu, Ma, and Mandal}{Belkin
  et~al.}{2019}]{belkin2019reconciling}
Belkin, M., D.~Hsu, S.~Ma, and S.~Mandal. 2019.
\newblock Reconciling modern machine-learning practice and the classical
  bias--variance trade-off.
\newblock {\em Proceedings of the National Academy of Sciences\/}~{\em
  116\/}(32): 15849--15854 .

\bibitem[\protect\citeauthoryear{Bird and Tobin}{Bird and
  Tobin}{2023}]{sep-natural-kinds}
Bird, A. and E.~Tobin. 2023.
\newblock {Natural Kinds}, In {\em The {Stanford} Encyclopedia of Philosophy\/}
  ({S}pring 2023 ed.).,  eds. Zalta, E.N. and U.~Nodelman. Metaphysics Research
  Lab, Stanford University.

\bibitem[\protect\citeauthoryear{Boge}{Boge}{2023}]{boge2023functional}
Boge, F.J. 2023.
\newblock Functional concept proxies and the actually smart hans problem:
  What’s special about deep neural networks in science.
\newblock {\em Synthese\/}~{\em 203\/}(1): 16 .

\bibitem[\protect\citeauthoryear{Bowers, Malhotra, Dujmovi{\'c}, Montero,
  Tsvetkov, Biscione, Puebla, Adolfi, Hummel, Heaton, et~al.}{Bowers
  et~al.}{2022}]{bowers2022deep}
Bowers, J.S., G.~Malhotra, M.~Dujmovi{\'c}, M.L. Montero, C.~Tsvetkov,
  V.~Biscione, G.~Puebla, F.~Adolfi, J.E. Hummel, R.F. Heaton, et~al. 2022.
\newblock Deep problems with neural network models of human vision.
\newblock {\em Behavioral and Brain Sciences\/}: 1--74 .

\bibitem[\protect\citeauthoryear{Brown, Man{\'e}, Roy, Abadi, and Gilmer}{Brown
  et~al.}{2017}]{brown2017adversarial}
Brown, T.B., D.~Man{\'e}, A.~Roy, M.~Abadi, and J.~Gilmer. 2017.
\newblock Adversarial patch.
\newblock {\em arXiv preprint arXiv:1712.09665\/} .

\bibitem[\protect\citeauthoryear{Buckner}{Buckner}{2018}]{buckner2018empiricism}
Buckner, C. 2018.
\newblock Empiricism without magic: Transformational abstraction in deep
  convolutional neural networks.
\newblock {\em Synthese\/}~{\em 195\/}(12): 5339--5372 .

\bibitem[\protect\citeauthoryear{Buckner}{Buckner}{2020}]{buckner2020understanding}
Buckner, C. 2020.
\newblock Understanding adversarial examples requires a theory of artefacts for
  deep learning.
\newblock {\em Nature Machine Intelligence\/}~{\em 2\/}(12): 731--736 .

\bibitem[\protect\citeauthoryear{Del~Pinal}{Del~Pinal}{2016}]{del2016prototypes}
Del~Pinal, G. 2016.
\newblock Prototypes as compositional components of concepts.
\newblock {\em Synthese\/}~193: 2899--2927 .

\bibitem[\protect\citeauthoryear{Donnelly and Roegiest}{Donnelly and
  Roegiest}{2019}]{donnelly2019interpretability}
Donnelly, J. and A.~Roegiest 2019.
\newblock On interpretability and feature representations: an analysis of the
  sentiment neuron.
\newblock In {\em Advances in Information Retrieval: 41st European Conference
  on IR Research, ECIR 2019, Cologne, Germany, April 14--18, 2019, Proceedings,
  Part I 41}, pp.\  795--802. Springer.

\bibitem[\protect\citeauthoryear{Duede}{Duede}{2023}]{duede2023representational}
Duede, E. 2023.
\newblock The representational status of deep learning models.
\newblock {\em arXiv preprint arXiv:2303.12032\/} .

\bibitem[\protect\citeauthoryear{Erickson, Korfiatis, Akkus, and
  Kline}{Erickson et~al.}{2017}]{erickson2017machine}
Erickson, B.J., P.~Korfiatis, Z.~Akkus, and T.L. Kline. 2017.
\newblock Machine learning for medical imaging.
\newblock {\em Radiographics\/}~{\em 37\/}(2): 505--515 .

\bibitem[\protect\citeauthoryear{Freiesleben}{Freiesleben}{2022}]{freiesleben2022intriguing}
Freiesleben, T. 2022.
\newblock The intriguing relation between counterfactual explanations and
  adversarial examples.
\newblock {\em Minds and Machines\/}~{\em 32\/}(1): 77--109 .

\bibitem[\protect\citeauthoryear{Freiesleben and Grote}{Freiesleben and
  Grote}{2023}]{freiesleben2023beyond}
Freiesleben, T. and T.~Grote. 2023.
\newblock Beyond generalization: a theory of robustness in machine learning.
\newblock {\em Synthese\/}~{\em 202\/}(4): 109 .

\bibitem[\protect\citeauthoryear{Freiesleben, K{\"o}nig, Molnar, and
  Tejero-Cantero}{Freiesleben et~al.}{2022}]{freiesleben2022scientific}
Freiesleben, T., G.~K{\"o}nig, C.~Molnar, and A.~Tejero-Cantero. 2022.
\newblock Scientific inference with interpretable machine learning: Analyzing
  models to learn about real-world phenomena.
\newblock {\em arXiv preprint arXiv:2206.05487\/} .

\bibitem[\protect\citeauthoryear{Frigg and Nguyen}{Frigg and
  Nguyen}{2021}]{sep-scientific-representation}
Frigg, R. and J.~Nguyen. 2021.
\newblock {Scientific Representation}, In {\em The {Stanford} Encyclopedia of
  Philosophy\/} ({W}inter 2021 ed.).,  ed. Zalta, E.N. Metaphysics Research
  Lab, Stanford University.

\bibitem[\protect\citeauthoryear{Gao, Cai, Fang, Li, Wang, Chen, Yu, Liu, Xu,
  Cui, et~al.}{Gao et~al.}{2020}]{gao2020machine}
Gao, Y., G.Y. Cai, W.~Fang, H.Y. Li, S.Y. Wang, L.~Chen, Y.~Yu, D.~Liu, S.~Xu,
  P.F. Cui, et~al. 2020.
\newblock Machine learning based early warning system enables accurate
  mortality risk prediction for covid-19.
\newblock {\em Nature communications\/}~{\em 11\/}(1): 5033 .

\bibitem[\protect\citeauthoryear{Geirhos, Jacobsen, Michaelis, Zemel, Brendel,
  Bethge, and Wichmann}{Geirhos et~al.}{2020}]{geirhos2020shortcut}
Geirhos, R., J.H. Jacobsen, C.~Michaelis, R.~Zemel, W.~Brendel, M.~Bethge, and
  F.A. Wichmann. 2020.
\newblock Shortcut learning in deep neural networks.
\newblock {\em Nature Machine Intelligence\/}~{\em 2\/}(11): 665--673 .

\bibitem[\protect\citeauthoryear{Geirhos, Rubisch, Michaelis, Bethge, Wichmann,
  and Brendel}{Geirhos et~al.}{2018}]{geirhos2018imagenet}
Geirhos, R., P.~Rubisch, C.~Michaelis, M.~Bethge, F.A. Wichmann, and W.~Brendel
  2018.
\newblock Imagenet-trained cnns are biased towards texture; increasing shape
  bias improves accuracy and robustness.
\newblock In {\em International Conference on Learning Representations}.

\bibitem[\protect\citeauthoryear{Goodfellow, Bengio, and Courville}{Goodfellow
  et~al.}{2016}]{goodfellow2016deep}
Goodfellow, I., Y.~Bengio, and A.~Courville. 2016.
\newblock {\em Deep learning}.
\newblock MIT press.

\bibitem[\protect\citeauthoryear{Goodfellow, Shlens, and Szegedy}{Goodfellow
  et~al.}{2014}]{goodfellow2014explaining}
Goodfellow, I.J., J.~Shlens, and C.~Szegedy. 2014.
\newblock Explaining and harnessing adversarial examples.
\newblock {\em arXiv preprint arXiv:1412.6572\/} .

\bibitem[\protect\citeauthoryear{Gruji{\v{c}}i{\'c}}{Gruji{\v{c}}i{\'c}}{2024}]{grujivcic2024deep}
Gruji{\v{c}}i{\'c}, B. 2024.
\newblock Deep convolutional neural networks are not mechanistic explanations
  of object recognition.
\newblock {\em Synthese\/}~{\em 203\/}(1): 1--28 .

\bibitem[\protect\citeauthoryear{Gurnee and Tegmark}{Gurnee and
  Tegmark}{2023}]{gurnee2023language}
Gurnee, W. and M.~Tegmark. 2023.
\newblock Language models represent space and time.
\newblock {\em arXiv preprint arXiv:2310.02207\/} .

\bibitem[\protect\citeauthoryear{Hampton}{Hampton}{2006}]{hampton2006concepts}
Hampton, J.A. 2006.
\newblock Concepts as prototypes.
\newblock {\em Psychology of learning and motivation\/}~46: 79--113 .

\bibitem[\protect\citeauthoryear{Hampton and Jönsson}{Hampton and
  Jönsson}{2012}]{hampton2012typicality}
Hampton, J.A. and M.L. Jönsson. 2012, 02.
\newblock {385 Typicality and Composition a Lity: the Logic of Combining Vague
  Concepts}, {\em {The Oxford Handbook of Compositionality}}. Oxford University
  Press.
\newblock \doi{10.1093/oxfordhb/9780199541072.013.0018}.

\bibitem[\protect\citeauthoryear{Hastie, Tibshirani, Friedman, and
  Friedman}{Hastie et~al.}{2009}]{hastie2009elements}
Hastie, T., R.~Tibshirani, J.H. Friedman, and J.H. Friedman. 2009.
\newblock {\em The elements of statistical learning: data mining, inference,
  and prediction}, Volume~2.
\newblock Springer.

\bibitem[\protect\citeauthoryear{Hornik, Stinchcombe, and White}{Hornik
  et~al.}{1989}]{hornik1989multilayer}
Hornik, K., M.~Stinchcombe, and H.~White. 1989.
\newblock Multilayer feedforward networks are universal approximators.
\newblock {\em Neural networks\/}~{\em 2\/}(5): 359--366 .

\bibitem[\protect\citeauthoryear{Ilyas, Santurkar, Tsipras, Engstrom, Tran, and
  Madry}{Ilyas et~al.}{2019}]{ilyas2019adversarial}
Ilyas, A., S.~Santurkar, D.~Tsipras, L.~Engstrom, B.~Tran, and A.~Madry. 2019.
\newblock Adversarial examples are not bugs, they are features.
\newblock {\em Advances in Neural Information Processing Systems\/}~32 .

\bibitem[\protect\citeauthoryear{Keil}{Keil}{1992}]{keil1992concepts}
Keil, F.C. 1992.
\newblock {\em Concepts, kinds, and cognitive development}.
\newblock MIT Press.

\bibitem[\protect\citeauthoryear{Kim, Wattenberg, Gilmer, Cai, Wexler, Viegas,
  et~al.}{Kim et~al.}{2018}]{kim2018interpretability}
Kim, B., M.~Wattenberg, J.~Gilmer, C.~Cai, J.~Wexler, F.~Viegas, et~al. 2018.
\newblock Interpretability beyond feature attribution: Quantitative testing
  with concept activation vectors (tcav).
\newblock In {\em International conference on machine learning}, pp.\
  2668--2677. PMLR.

\bibitem[\protect\citeauthoryear{Kim, Lee, and Choo}{Kim
  et~al.}{2021}]{kim2021biaswap}
Kim, E., J.~Lee, and J.~Choo 2021.
\newblock Biaswap: Removing dataset bias with bias-tailored swapping
  augmentation.
\newblock In {\em Proceedings of the IEEE/CVF International Conference on
  Computer Vision}, pp.\  14992--15001.

\bibitem[\protect\citeauthoryear{Kingma and Ba}{Kingma and
  Ba}{2014}]{kingma2014adam}
Kingma, D.P. and J.~Ba. 2014.
\newblock Adam: A method for stochastic optimization.
\newblock {\em arXiv preprint arXiv:1412.6980\/} .

\bibitem[\protect\citeauthoryear{Koh, Nguyen, Tang, Mussmann, Pierson, Kim, and
  Liang}{Koh et~al.}{2020}]{koh2020concept}
Koh, P.W., T.~Nguyen, Y.S. Tang, S.~Mussmann, E.~Pierson, B.~Kim, and P.~Liang
  2020.
\newblock Concept bottleneck models.
\newblock In {\em International conference on machine learning}, pp.\
  5338--5348. PMLR.

\bibitem[\protect\citeauthoryear{K{\"o}nig, Freiesleben, and
  Grosse-Wentrup}{K{\"o}nig et~al.}{2023}]{konig2023improvement}
K{\"o}nig, G., T.~Freiesleben, and M.~Grosse-Wentrup 2023.
\newblock Improvement-focused causal recourse (icr).
\newblock In {\em Proceedings of the AAAI Conference on Artificial
  Intelligence}, Volume~37, pp.\  11847--11855.

\bibitem[\protect\citeauthoryear{Kornblith, Shlens, and Le}{Kornblith
  et~al.}{2019}]{kornblith2019better}
Kornblith, S., J.~Shlens, and Q.V. Le 2019.
\newblock Do better imagenet models transfer better?
\newblock In {\em Proceedings of the IEEE/CVF conference on computer vision and
  pattern recognition}, pp.\  2661--2671.

\bibitem[\protect\citeauthoryear{Lalumera}{Lalumera}{2010}]{lalumera2010concepts}
Lalumera, E. 2010.
\newblock Concepts are a functional kind.
\newblock {\em Behavioral and Brain Sciences\/}~{\em 33\/}(2-3): 217 .

\bibitem[\protect\citeauthoryear{Leavitt and Morcos}{Leavitt and
  Morcos}{2020a}]{leavitt2020towards}
Leavitt, M.L. and A.~Morcos. 2020a.
\newblock Towards falsifiable interpretability research.
\newblock {\em arXiv preprint arXiv:2010.12016\/} .

\bibitem[\protect\citeauthoryear{Leavitt and Morcos}{Leavitt and
  Morcos}{2020b}]{leavitt2020relationship}
Leavitt, M.L. and A.S. Morcos. 2020b.
\newblock On the relationship between class selectivity, dimensionality, and
  robustness.
\newblock {\em arXiv preprint arXiv:2007.04440\/} .

\bibitem[\protect\citeauthoryear{Leavitt and Morcos}{Leavitt and
  Morcos}{2020c}]{leavitt2020selectivity}
Leavitt, M.L. and A.S. Morcos 2020c.
\newblock Selectivity considered harmful: evaluating the causal impact of class
  selectivity in dnns.
\newblock In {\em International Conference on Learning Representations}.

\bibitem[\protect\citeauthoryear{LeCun, Bengio, and Hinton}{LeCun
  et~al.}{2015}]{lecun2015deep}
LeCun, Y., Y.~Bengio, and G.~Hinton. 2015.
\newblock Deep learning.
\newblock {\em nature\/}~{\em 521\/}(7553): 436--444 .

\bibitem[\protect\citeauthoryear{Lipton}{Lipton}{2018}]{lipton2018mythos}
Lipton, Z.C. 2018.
\newblock The mythos of model interpretability: In machine learning, the
  concept of interpretability is both important and slippery.
\newblock {\em Queue\/}~{\em 16\/}(3): 31--57 .

\bibitem[\protect\citeauthoryear{L{\'o}pez-Rubio}{L{\'o}pez-Rubio}{2021}]{lopez2021throwing}
L{\'o}pez-Rubio, E. 2021.
\newblock Throwing light on black boxes: emergence of visual categories from
  deep learning.
\newblock {\em Synthese\/}~{\em 198\/}(10): 10021--10041 .

\bibitem[\protect\citeauthoryear{Lu, Pu, Wang, Hu, and Wang}{Lu
  et~al.}{2017}]{lu2017expressive}
Lu, Z., H.~Pu, F.~Wang, Z.~Hu, and L.~Wang. 2017.
\newblock The expressive power of neural networks: A view from the width.
\newblock {\em Advances in neural information processing systems\/}~30 .

\bibitem[\protect\citeauthoryear{Marblestone, Wayne, and Kording}{Marblestone
  et~al.}{2016}]{marblestone2016toward}
Marblestone, A.H., G.~Wayne, and K.P. Kording. 2016.
\newblock Toward an integration of deep learning and neuroscience.
\newblock {\em Frontiers in computational neuroscience\/}~10: 94 .

\bibitem[\protect\citeauthoryear{McKenna, Li, Cheng, Hosseini, Johnson, and
  Steedman}{McKenna et~al.}{2023}]{mckenna2023sources}
McKenna, N., T.~Li, L.~Cheng, M.J. Hosseini, M.~Johnson, and M.~Steedman. 2023.
\newblock Sources of hallucination by large language models on inference tasks.
\newblock {\em arXiv preprint arXiv:2305.14552\/} .

\bibitem[\protect\citeauthoryear{Meng, Bau, Andonian, and Belinkov}{Meng
  et~al.}{2022}]{meng2022locating}
Meng, K., D.~Bau, A.~Andonian, and Y.~Belinkov. 2022.
\newblock Locating and editing factual associations in gpt.
\newblock {\em Advances in Neural Information Processing Systems\/}~35:
  17359--17372 .

\bibitem[\protect\citeauthoryear{Morcos, Barrett, Rabinowitz, and
  Botvinick}{Morcos et~al.}{2018}]{morcos2018importance}
Morcos, A.S., D.G. Barrett, N.C. Rabinowitz, and M.~Botvinick 2018.
\newblock On the importance of single directions for generalization.
\newblock In {\em International Conference on Learning Representations}.

\bibitem[\protect\citeauthoryear{Mukhlif, Al-Khateeb, and Mohammed}{Mukhlif
  et~al.}{2023}]{mukhlif2023incorporating}
Mukhlif, A.A., B.~Al-Khateeb, and M.A. Mohammed. 2023.
\newblock Incorporating a novel dual transfer learning approach for medical
  images.
\newblock {\em Sensors\/}~{\em 23\/}(2): 570 .

\bibitem[\protect\citeauthoryear{Oikarinen, Srinivasan, Meisner, Hyman, Parmar,
  Fanucci-Kiss, Desimone, Landman, and Feng}{Oikarinen
  et~al.}{2019}]{oikarinen2019deep}
Oikarinen, T., K.~Srinivasan, O.~Meisner, J.B. Hyman, S.~Parmar,
  A.~Fanucci-Kiss, R.~Desimone, R.~Landman, and G.~Feng. 2019.
\newblock Deep convolutional network for animal sound classification and source
  attribution using dual audio recordings.
\newblock {\em The Journal of the Acoustical Society of America\/}~{\em
  145\/}(2): 654--662 .

\bibitem[\protect\citeauthoryear{Olah, Cammarata, Schubert, Goh, Petrov, and
  Carter}{Olah et~al.}{2020}]{olah2020zoom}
Olah, C., N.~Cammarata, L.~Schubert, G.~Goh, M.~Petrov, and S.~Carter. 2020.
\newblock Zoom in: An introduction to circuits.
\newblock {\em Distill\/}~{\em 5\/}(3): e00024--001 .

\bibitem[\protect\citeauthoryear{Olah, Mordvintsev, and Schubert}{Olah
  et~al.}{2017}]{olah2017feature}
Olah, C., A.~Mordvintsev, and L.~Schubert. 2017.
\newblock Feature visualization.
\newblock {\em Distill\/}~{\em 2\/}(11): e7 .

\bibitem[\protect\citeauthoryear{Peacocke}{Peacocke}{1992}]{peacocke1992study}
Peacocke, C. 1992.
\newblock {\em A study of concepts.}
\newblock MIT Press.

\bibitem[\protect\citeauthoryear{Popper}{Popper}{2005}]{popper2005logic}
Popper, K. 2005.
\newblock {\em The logic of scientific discovery}.
\newblock Routledge.

\bibitem[\protect\citeauthoryear{Quine}{Quine}{1969}]{quine1969ontological}
Quine, W.V. 1969.
\newblock {\em Ontological relativity and other essays}.
\newblock Columbia University Press.

\bibitem[\protect\citeauthoryear{Radford, Jozefowicz, and Sutskever}{Radford
  et~al.}{2017}]{radford2017learning}
Radford, A., R.~Jozefowicz, and I.~Sutskever. 2017.
\newblock Learning to generate reviews and discovering sentiment.
\newblock {\em arXiv preprint arXiv:1704.01444\/} .

\bibitem[\protect\citeauthoryear{R{\"a}z}{R{\"a}z}{2023}]{raz2023methods}
R{\"a}z, T. 2023.
\newblock Methods for identifying emergent concepts in deep neural networks.
\newblock {\em Patterns\/}~{\em 4\/}(6) .

\bibitem[\protect\citeauthoryear{Ren, Li, Ren, Song, Xu, Deng, and Wang}{Ren
  et~al.}{2021}]{ren2021deep}
Ren, X., X.~Li, K.~Ren, J.~Song, Z.~Xu, K.~Deng, and X.~Wang. 2021.
\newblock Deep learning-based weather prediction: a survey.
\newblock {\em Big Data Research\/}~23: 100178 .

\bibitem[\protect\citeauthoryear{Ribeiro, Singh, and Guestrin}{Ribeiro
  et~al.}{2016}]{ribeiro2016should}
Ribeiro, M.T., S.~Singh, and C.~Guestrin 2016.
\newblock "why should i trust you?" explaining the predictions of any
  classifier.
\newblock In {\em Proceedings of the 22nd ACM SIGKDD international conference
  on knowledge discovery and data mining}, pp.\  1135--1144.

\bibitem[\protect\citeauthoryear{Rolnick and Tegmark}{Rolnick and
  Tegmark}{2018}]{rolnick2017power}
Rolnick, D. and M.~Tegmark 2018.
\newblock The power of deeper networks for expressing natural functions.
\newblock In {\em International Conference on Learning Representations}.

\bibitem[\protect\citeauthoryear{Sch{\"o}lkopf, Locatello, Bauer, Ke,
  Kalchbrenner, Goyal, and Bengio}{Sch{\"o}lkopf
  et~al.}{2021}]{scholkopf2021toward}
Sch{\"o}lkopf, B., F.~Locatello, S.~Bauer, N.R. Ke, N.~Kalchbrenner, A.~Goyal,
  and Y.~Bengio. 2021.
\newblock Toward causal representation learning.
\newblock {\em Proceedings of the IEEE\/}~{\em 109\/}(5): 612--634 .

\bibitem[\protect\citeauthoryear{Shermin, Teng, Murshed, Lu, Sohel, and
  Paul}{Shermin et~al.}{2019}]{shermin2019enhanced}
Shermin, T., S.W. Teng, M.~Murshed, G.~Lu, F.~Sohel, and M.~Paul 2019.
\newblock Enhanced transfer learning with imagenet trained classification
  layer.
\newblock In {\em Image and Video Technology: 9th Pacific-Rim Symposium, PSIVT
  2019, Sydney, NSW, Australia, November 18--22, 2019, Proceedings 9}, pp.\
  142--155. Springer.

\bibitem[\protect\citeauthoryear{Sterkenburg}{Sterkenburg}{2023}]{sterkenburg2023statistical}
Sterkenburg, T.F. 2023.
\newblock Statistical learning theory and occam's razor: The argument from
  empirical risk minimization.
\newblock {\em arXiv preprint arXiv:2312.13842\/} .

\bibitem[\protect\citeauthoryear{Szegedy, Zaremba, Sutskever, Bruna, Erhan,
  Goodfellow, and Fergus}{Szegedy et~al.}{2013}]{szegedy2013intriguing}
Szegedy, C., W.~Zaremba, I.~Sutskever, J.~Bruna, D.~Erhan, I.~Goodfellow, and
  R.~Fergus. 2013.
\newblock Intriguing properties of neural networks.
\newblock {\em arXiv preprint arXiv:1312.6199\/} .

\bibitem[\protect\citeauthoryear{Tan, Sun, Kong, Zhang, Yang, and Liu}{Tan
  et~al.}{2018}]{tan2018survey}
Tan, C., F.~Sun, T.~Kong, W.~Zhang, C.~Yang, and C.~Liu 2018.
\newblock A survey on deep transfer learning.
\newblock In {\em Artificial Neural Networks and Machine Learning--ICANN 2018:
  27th International Conference on Artificial Neural Networks, Rhodes, Greece,
  October 4-7, 2018, Proceedings, Part III 27}, pp.\  270--279. Springer.

\bibitem[\protect\citeauthoryear{Vapnik}{Vapnik}{1999}]{vapnik1999nature}
Vapnik, V. 1999.
\newblock {\em The nature of statistical learning theory}.
\newblock Springer science \& business media.

\bibitem[\protect\citeauthoryear{Vaswani, Shazeer, Parmar, Uszkoreit, Jones,
  Gomez, Kaiser, and Polosukhin}{Vaswani et~al.}{2017}]{vaswani2017attention}
Vaswani, A., N.~Shazeer, N.~Parmar, J.~Uszkoreit, L.~Jones, A.N. Gomez,
  {\L}.~Kaiser, and I.~Polosukhin. 2017.
\newblock Attention is all you need.
\newblock {\em Advances in neural information processing systems\/}~30 .

\bibitem[\protect\citeauthoryear{Von~Luxburg and Sch{\"o}lkopf}{Von~Luxburg and
  Sch{\"o}lkopf}{2011}]{von2011statistical}
Von~Luxburg, U. and B.~Sch{\"o}lkopf. 2011.
\newblock Statistical learning theory: Models, concepts, and results, {\em
  Handbook of the History of Logic}, Volume~10,  651--706. Elsevier.

\bibitem[\protect\citeauthoryear{Wang and Gong}{Wang and
  Gong}{2022}]{wang2022adversarial}
Wang, S. and Y.~Gong. 2022.
\newblock Adversarial example detection based on saliency map features.
\newblock {\em Applied Intelligence\/}: 1--14 .

\bibitem[\protect\citeauthoryear{Watson}{Watson}{2023}]{watson2023philosophy}
Watson, D.S. 2023.
\newblock On the philosophy of unsupervised learning.
\newblock {\em Philosophy \& Technology\/}~{\em 36\/}(2): 28 .

\bibitem[\protect\citeauthoryear{Williams}{Williams}{2018}]{williams2018predictive}
Williams, D. 2018.
\newblock Predictive processing and the representation wars.
\newblock {\em Minds and Machines\/}~{\em 28\/}(1): 141--172 .

\bibitem[\protect\citeauthoryear{Wittgenstein}{Wittgenstein}{2010}]{wittgenstein2010philosophical}
Wittgenstein, L. 2010.
\newblock {\em Philosophical investigations}.
\newblock John Wiley \& Sons.

\bibitem[\protect\citeauthoryear{Yang, Li, Wang, Dong, Wang, and Tang}{Yang
  et~al.}{2020}]{yang2020using}
Yang, J., S.~Li, Z.~Wang, H.~Dong, J.~Wang, and S.~Tang. 2020.
\newblock Using deep learning to detect defects in manufacturing: a
  comprehensive survey and current challenges.
\newblock {\em Materials\/}~{\em 13\/}(24): 5755 .

\bibitem[\protect\citeauthoryear{Yee}{Yee}{2019}]{yee2019abstraction}
Yee, E. 2019.
\newblock Abstraction and concepts: when, how, where, what and why?
\newblock {\em Language, Cognition and Neuroscience\/}~{\em 34\/}(10):
  1257--1265 .

\bibitem[\protect\citeauthoryear{Zhang, Bengio, Hardt, Recht, and
  Vinyals}{Zhang et~al.}{2021}]{zhang2021understanding}
Zhang, C., S.~Bengio, M.~Hardt, B.~Recht, and O.~Vinyals. 2021.
\newblock Understanding deep learning (still) requires rethinking
  generalization.
\newblock {\em Communications of the ACM\/}~{\em 64\/}(3): 107--115 .

\bibitem[\protect\citeauthoryear{Zhang, Yao, Sun, and Tay}{Zhang
  et~al.}{2019}]{zhang2019deep}
Zhang, S., L.~Yao, A.~Sun, and Y.~Tay. 2019.
\newblock Deep learning based recommender system: A survey and new
  perspectives.
\newblock {\em ACM computing surveys (CSUR)\/}~{\em 52\/}(1): 1--38 .

\bibitem[\protect\citeauthoryear{Zhong, Ding, Liu, Du, and Tao}{Zhong
  et~al.}{2023}]{zhong2023can}
Zhong, Q., L.~Ding, J.~Liu, B.~Du, and D.~Tao. 2023.
\newblock Can chatgpt understand too? a comparative study on chatgpt and
  fine-tuned bert.
\newblock {\em arXiv preprint arXiv:2302.10198\/} .

\bibitem[\protect\citeauthoryear{Zhou, Sun, Bau, and Torralba}{Zhou
  et~al.}{2018}]{zhou2018revisiting}
Zhou, B., Y.~Sun, D.~Bau, and A.~Torralba. 2018.
\newblock Revisiting the importance of individual units in cnns via ablation.
\newblock {\em arXiv preprint arXiv:1806.02891\/} .

\end{thebibliography}

\end{document}